\documentclass[letterpaper, 10 pt, journal, twoside]{IEEEtran}
\usepackage{graphicx}                
\usepackage{balance}
\usepackage{amsmath,amsfonts,amssymb,mathtools}     
\usepackage[dvipsnames]{xcolor}
\usepackage{array}
\usepackage{xfrac} 
\newtheorem{proposition}{Proposition}

\newtheorem{remark}{Remark}

\newcommand{\dom}{\operatorname{dom}}
\makeatletter
\def\maketag@@@#1{\hbox{\m@th\normalfont\normalsize#1}}
\makeatother

\begin{document}

\title{A Hybrid Dynamical Modeling Framework for Shape Memory Alloy Wire Actuated Structures}

\author{Michele A. Mandolino$^{1}$, Francesco Ferrante$^{2}$~\IEEEmembership{Member,~IEEE}, and Gianluca Rizzello$^{1}$~\IEEEmembership{Member,~IEEE}%

\thanks{Manuscript received: October, 15, 2020; Revised December, 19, 2020; Accepted February, 20, 2021.}

\thanks{This paper was recommended for publication by Editor Cecilia Laschi upon evaluation of the Associate Editor and Reviewers' comments.
}
\thanks{$^{1}$Michele A. Mandolino and Gianluca Rizzello are with the Department of Systems Engineering, Department of Material Science and Engineering, Saarland University, 66123 Saarbr\"ucken, Germany
        {\tt\footnotesize \{michele.mandolino, gianluca.rizzello\}@imsl.uni-saarland.de}}%
\thanks{$^{2} $Francesco Ferrante is with University of Grenoble Alpes, CNRS, Grenoble INP, GIPSA-lab, 38000 Grenoble, France
        {\tt\footnotesize francesco.ferrante@univ-grenoble-alpes.fr}}%
\thanks{Digital Object Identifier (DOI): see top of this page.}
}

\markboth{IEEE Robotics and Automation Letters. Preprint Version. Accepted March, 2021}
{Mandolino \MakeLowercase{\textit{et al.}}: A Hybrid Dynamical Modeling Framework For Shape Memory Alloy Wire Actuated Structures} 

\maketitle

\begin{abstract}
In this paper, a hybrid model for single-crystal Shape Memory Alloy (SMA) wire actuators is presented. The result is based on a mathematical reformulation of the M\"{u}ller-Achenbach-Seelecke (MAS) model, which provides an accurate and interconnection-oriented description of the SMA hysteretic response. The strong nonlinearity and high numerical stiffness of the MAS model, however, hinder its practical use for simulation and control of complex SMA-driven systems.
The main idea behind the hybrid reformulation is based on dividing the mechanical hysteresis of the SMA into five operating modes, each one representing a different physical state of the material.
By properly deriving the switching conditions among those modes in a physically-consistent way, the MAS model is effectively reformulated within a hybrid dynamical setting.
The main advantage of the hybrid reformulation is the possibility of describing the material dynamics with a simplified set of state equations while maintaining all benefits of the physics-based description offered by the MAS model
After describing the novel approach, simulation studies are conducted on a flexible robotic module actuated by protagonist-antagonist SMA wires. Through comparative numerical analysis, it is shown how the hybrid model provides the same accuracy as the MAS model while saving up to 80\% of the simulation time.
Moreover, the new modeling framework opens up the possibility of addressing SMA control from a hybrid systems perspective.
\end{abstract}


\IEEEpeerreviewmaketitle

\section{Introduction}

\IEEEPARstart{A}{s} a result of their unique flexibility and scalability, continuum robots have found applications in challenging areas such as medical \cite{Burgner-Kahrs:2015}, industrial maintenance \cite{Dong:2019}, and inspection \cite{Wang:2021}. Mechanical structures conventionally adopted for the realization of continuum robots allow them to reach specific positions without the aid of joints or rigid links. The actuators commonly adopted in those applications are either based on pneumatic systems \cite{Hofer:2018} or motorized tendons \cite{Gao:2019}. Despite their popularity, those actuation solutions are often noisy, not efficient, and bulky, with an overall size which is significantly larger than the true dimension of the active part. 

It is remarked how Shape Memory Alloy (SMA) wires represent a promising technology to improve the performance and further the miniaturization of continuum robots. A SMA consists of a metal alloy which contracts when heated via an electric current. This phenomenon is generated by a phase transformation in the material microstructure, which results into macroscopic changes in shape up to 4-8\%. After removing the electric current, the initial shape of the wire can be recovered by applying an external force (provided, e.g., by a spring load or by another SMA wire). Features such as high energy density, flexibility, and self-sensing operations make SMA wires particularly attractive for many application fields, including bioinspired robots \cite{Laschi:2012}, endoscopes \cite{Giataganas:2011}, and artificial hands \cite{Simone:2017}. However, the temperature- and rate-dependent hysteretic response of the material makes the design, modeling, and control of SMA systems a highly challenging task. This issue is even more critical in case multiple SMA wires are used to activate a complex mechanical structure, thus resulting in a strongly nonlinear dynamic system \cite{Crews:2012}, \cite{Liu:2018}, \cite{Ge:2007}. The development of models and simulation tools, which account for the physical coupling between actuator and structure in a numerically efficient way, represents a fundamental step towards the design and control of high-performance SMA robots.

With the aim of developing improved numerical tools for SMA systems, in this paper we present a novel hybrid model for single-crystal SMA wire actuators. The proposed hybrid reformulation is grounded on the SMA model originally developed by M{\"u}ller-Achenbach-Seelecke (MAS) which, in turn, is based on a statistical thermodynamic framework \cite{Seelecke:2004}, \cite{Ballew:2019}. Due to its physics-based nature, the MAS model allows to effectively describe the hysteretic behavior of SMA wire actuators under different operating conditions and thermo-mechanical loads. The high level of detail of the MAS model, however, results in strong nonlinearities which affect the simulation time and complicate the design of control systems. By properly exploiting some structural properties of the MAS model, the original system equations are reformulated within the hybrid systems framework in \cite{Goebel:2012}. The new hybrid model enables to describe the same input-output behavior of the original MAS one by means of a simplified set of equations, which rely on a reduced number of continuous state variables. In this way, all the physical features of the original model can be accounted for in a more numerically efficient way. At the same time, the hybrid formalism opens up the possibility of analyzing and controlling complex SMA systems based on powerful analytical tools \cite{Goebel:2012}.
In our previous work, we have shown for the first time the effectiveness of the hybrid framework in modeling hysteretic SMA systems \cite{Mandolino:2020}. The result, however, is only valid for a specific type of SMA-spring actuator. Since the introduction of the spring element drastically modifies the constitutive equations of the SMA wire, the developed model cannot be used to describe more complex types of SMA systems. In this paper, we extend the results in \cite{Mandolino:2020} by developing:
\begin{itemize}
    \item An analytical characterization of the different modes occurring in the hysteresis of single-crystal SMAs;
    \item A new hybrid formulation for the model of a generic single-crystal SMA wire, in a form which is suitable for port-interconnection with a mechanical structure;
    \item An experimental validation of the SMA hybrid model;
    \item A simulation study of a flexible robotic structure actuated by bundles of SMA wires, in which the performance of both MAS and hybrid models are compared.
\end{itemize}

The remainder of this paper is organized as follows. In Section~\ref{sec:2}, a motivating example of SMA continuum robot is presented. Section~\ref{sec:3} provides a formal and exhaustive description of the new hybrid model, while parameter identification and simulation studies are reported in Section~\ref{sec:4}.
Finally, concluding remarks are discussed in Section~\ref{sec:5}.

\section{Motivation: SMA Wire Driven Flexible Robots}
\label{sec:2}

\begin{figure}  \vspace{0.1cm}
    \begin{center}
        \includegraphics[width=8.0cm]{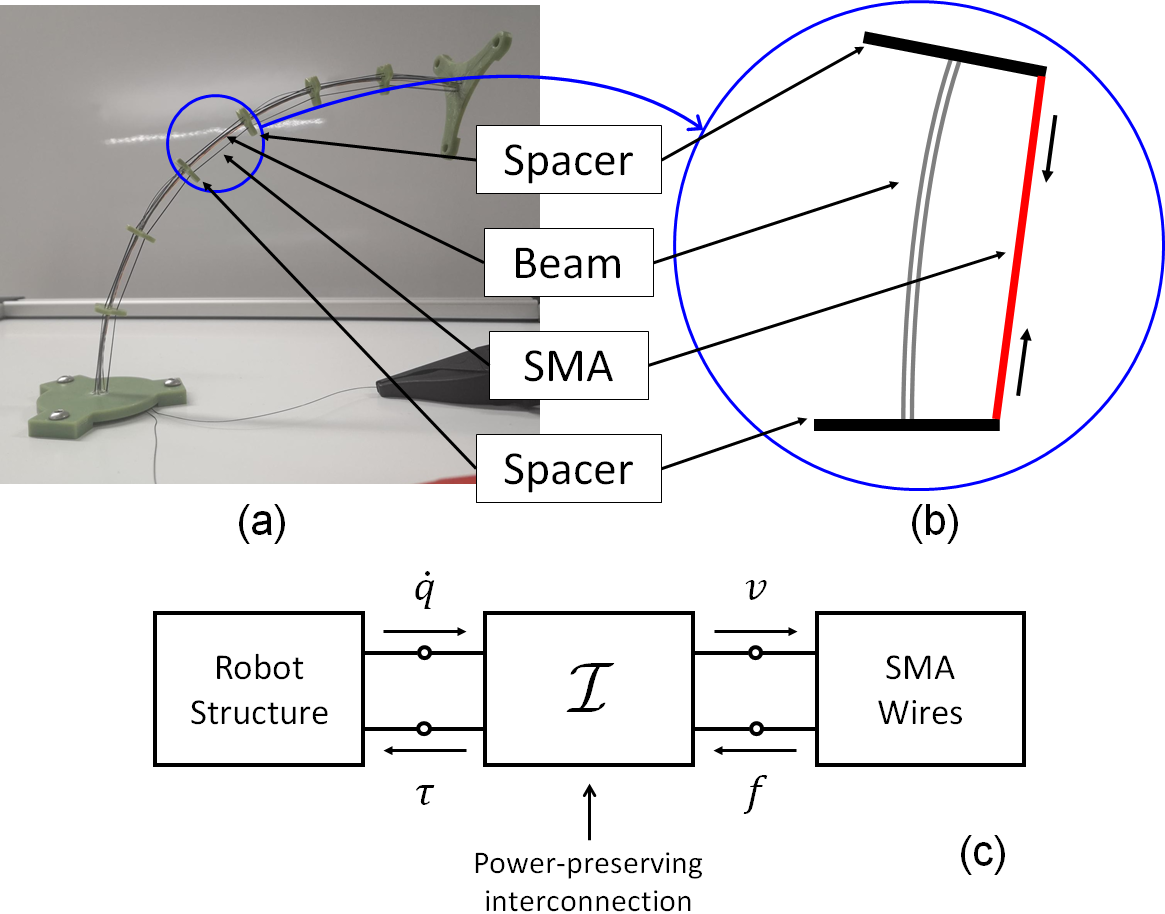}
        \caption{Example of SMA continuum robot \cite{Yannik:2020} (a) and its graphical representation with a flexible link between two rigid spacer (b). Port-based model of the structure based on the power-preserving interconnection $\mathcal{I}$ (c).}
        \label{fig:coupling}
    \end{center}
\end{figure}

An example of SMA continuum robot is shown in Fig.~\ref{fig:coupling}(a) \cite{Yannik:2020}. It is based on a serial connection of several bendable modules, each one consisting of a flexible backbone connecting two rigid spacers. Three bundles of SMA wires, equally spaced around the rigid spacer, are connected in parallel to the backbone, working as actuator elements. 
When the SMAs are inactive, the structure is in a neutral position. By activating the wires, the resulting contraction in length generates an asymmetry in the forces applied to the plate and, in turn, a bending of the beam in the direction of the actuated bundle, see Fig.~\ref{fig:coupling}(b). By stacking many of those modules, a flexible robotic arm capable of large bending angles can be obtained. 

To consistently model a continuum robot such as the one in Fig.~\ref{fig:coupling}(a), an interconnection-based viewpoint can be effectively adopted. A schematic depiction of such an approach is shown in Fig.~\ref{fig:coupling}(c). The block \emph{Robot Structure} describes the model of the robot itself (i.e., the combined flexible backbone and spacers). Since those model are generally obtained via Newtonian or Euler-Lagrange approaches, their causal representation normally comes in the form of a mechanical admittance (force-input, velocity-output) \cite{Duindam:2009,Rucker:2011,Falkenhahn:2015,Rizzello:2017}. To implement actuation, the structure needs to be coupled with a model of the \emph{SMA Wires}. This is possible via a power-preserving interconnection $\mathcal{I}$, which acts as a force/velocity transformer among the two subsystems \cite{Duindam:2009}. In order to effectively implement this type of architecture without violating causality, the SMA model needs to be provided in the form of a mechanical impedance (velocity-input, force-output). Moreover, if the SMA model satisfies energy consistency properties (i.e., passivity), stability of the interconnection can be guaranteed for any passive mechanical structure \cite{Duindam:2009}. Those features will be accounted for when developing the hybrid SMA model.

\section{Single-crystal SMA Model}
\label{sec:3}

In this section, the MAS model for single-crystal SMA wires is first summarized. Subsequently, an exhaustive description of the hybrid reformulation is proposed.

\subsection{MAS Model}

The MAS model provides a description of single-crystal SMA material based on a general mesoscopic free-energy framework \cite{Furst:2012}. A SMA consist of a metal alloy which contracts due to a phase transformation among lattice variants when heated. In the specific case of a SMA wire actuator we consider only two variants denoted, as austenite (A) and martensite-plus (M). 
According to MAS, the macroscopic stress-strain relationship of the material can be expressed as follows
\begin{equation} \label{eq:MAS3}
    \sigma = \sigma(\varepsilon,x_M) = \frac{\varepsilon-\varepsilon_T x_M}{E_M^{-1}{x_M} + E_A^{-1}(1-x_M)},
\end{equation}
where $\varepsilon$ is the SMA strain, $\sigma$ is the SMA stress, $x_M$ is the martensitic variant phase fraction, while $E_A$, $E_M$, and $\varepsilon_T$ are constant constitutive material parameters and represent the austenite Young's modulus, martensite Young's modulus, and transformation strain, respectively. 
When the phase fraction $x_M$ varies from 0 (full austenite) to 1 (full martensite), the angular coefficient of the hysteretic curve, equivalent to Young's modulus calculated at the operating point, changes from $E_A$ to $E_M$ and vice versa. As we will see in the sequel, the dependence of the material stress of the phase fraction $x_M$ introduces a temperature-dependent hysteretic behavior.

\begin{figure} \vspace{0.1cm}
    \begin{center}
        \includegraphics[width=8.8cm]{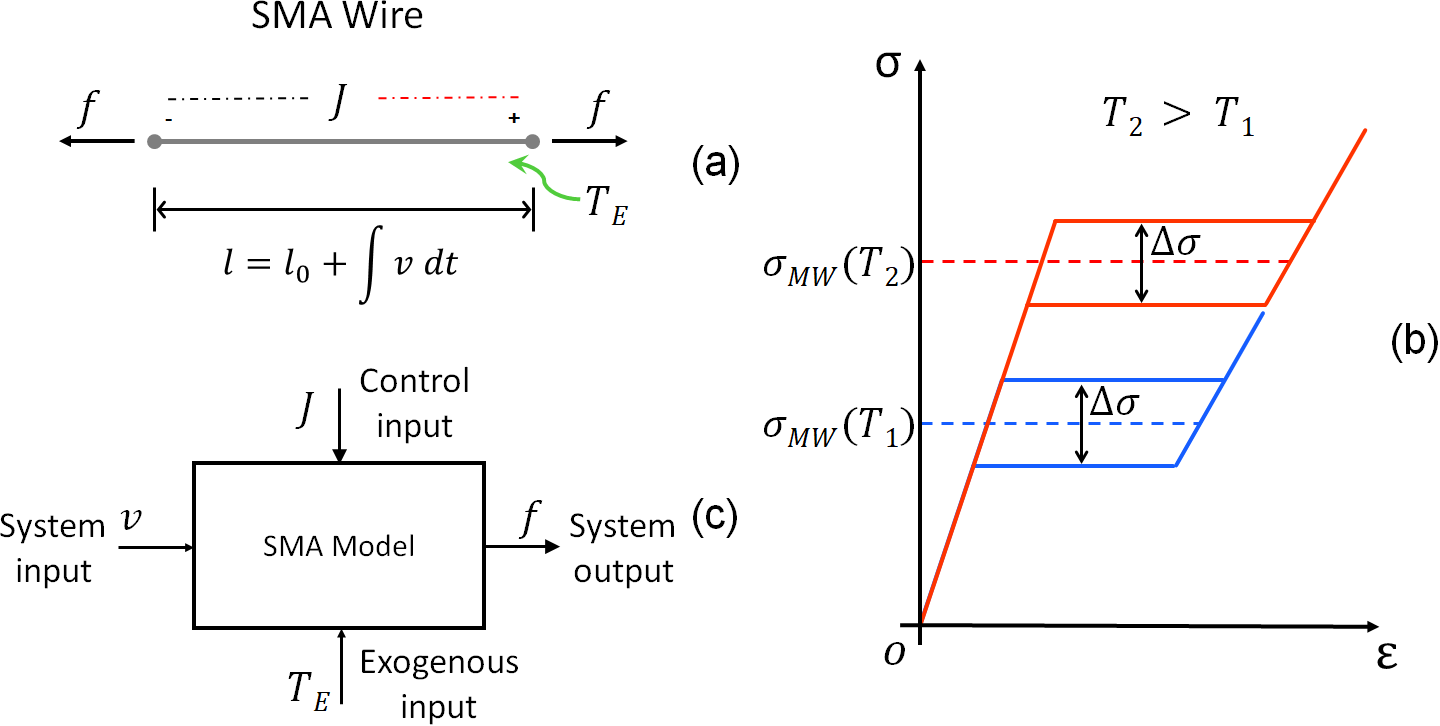}
        \caption{Inputs and outputs for a generic SMA wire (a), SMA hysteresis at two different temperatures (b), and SMA model block diagram (c).}
        \label{fig:SMA1}
    \end{center}
\end{figure}

Material stress and strain can be related to the wire force $f$, length $l$, and deformation rate $\textit{v}$ (cf. Fig.~\ref{fig:SMA1}(a)) by means of the following equations
\begin{align} \label{eq:MAS4}
    f &=\pi r_0^2 \sigma,\\
    l &= l_0 (1 + \varepsilon),\\
    \label{eq:MAS4a}
    \textit{v} &= \dot{l} = l_0 \dot{\varepsilon},
\end{align}
where $r_0$ and $l_0$ are the radius and the length of the undeformed SMA wire, respectively.

To describe the dynamic evolution of $x_M$, the following equation can be derived through statistical thermodynamics
\begin{equation} \label{eq:MAS5}
    \dot{x}_M = -p_{MA}x_M + p_{AM}\left(1-x_M\right).
\end{equation}
The generic transition probability $p_{\alpha \beta}$ from phase $\alpha$ to phase $\beta$ can be computed as follows
\begin{equation} \label{eq:MAS6}
    p_{\alpha \beta} = p_{\alpha \beta}(\sigma,T) = \omega_x e^{-\frac{V_L}{k_B T}\Delta g_{\alpha \beta}(\sigma,T)},
\end{equation}
where $\omega_x$ is a natural frequency associated to thermal activation, $V_L$ is the volume of a mesoscopic crystal layer, $k_B$ is the Boltzmann constant, $T$ is the SMA temperature, and $\Delta g_{\alpha \beta}(\sigma,T)$ is the Gibbs free-energy barrier of the phase transformation.
The energy barriers $\Delta g_{\alpha \beta}$ depend in a complex mathematical way on the transformation stresses of austenite and martensite, denoted as $\sigma_A$ and $\sigma_M$ respectively (details are omitted for conciseness, the reader may refer to \cite{Seelecke:2004,Ballew:2019} for details). Such quantities are given by
\begin{align} 
\label{eq:MAS7}
    \sigma_A &= \sigma_A(T) = \sigma_{MW}(T) + 0.5\Delta\sigma, \\
    \label{eq:MAS8}
    \sigma_M &= \sigma_M(T) = \sigma_{MW}(T) - 0.5\Delta\sigma,
\end{align}
where $\Delta\sigma$ is the size of the $\sigma$-$\varepsilon$ hysteresis, while
\begin{equation} \label{eq:MAS8a}
    \sigma_{MW}(T) = \sigma_{MW}(T_0) + \sigma_T\left( T - T_0 \right),
\end{equation}
for some constant parameters $T_0$ and $\sigma_T$. The physical meaning of $\Delta\sigma$ and $\sigma_{MW}$ is shown in Fig.~\ref{fig:SMA1}(b).

Finally, the temperature evolution of the SMA can be derived from the following internal energy balance equation
\begin{equation} \label{eq:MAS9}
    \Omega \rho_V c_V \dot{T} = - \lambda A_S (T - T_E) + J + \dot{L},
\end{equation}
where $\Omega = \pi r_0^2l_0$ is the SMA material volume, $\rho_V$ is the SMA density, $c_V$ is the SMA specific heat, $\lambda$ is the convective cooling coefficient between SMA wire and environment, $A_s=2\pi r_0l_0$ is the lateral surface area of the wire, $T_E$ is the environmental temperature, $J$ is the Joule heating produced by an electric current, and $\dot{L}$ is the internal latent heat production due to the phase transformation.
In this work, the latent heat $\dot{L}$ in (\ref{eq:MAS9}) is modeled according to the following
\begin{equation} \label{eq:L3}
    \dot{L} = L_{x_M}\dot{x}_M + L_T\dot{T},
\end{equation}
where functions $L_{x_M}$ and $L_{x_T}$ are given by
\begin{align} 
\label{eq:L3a}
    L_{x_M} & = \Omega \Big(T \Gamma_T(T) + \Gamma(\sigma) - \Gamma(\sigma_{MW}(T))\Big), \\
\label{eq:L3b}
    L_T & = \Omega T \Gamma_{TT}(T)(x_M - 1),
\end{align}
with
\begin{align} 
    \label{eq:L2a}
    \Gamma(\sigma) & = \left( E_M^{-1} - E_A^{-1} \right)0.5\sigma^2+\varepsilon_T\sigma, \\
    \label{eq:L2b}
    \Gamma_T(T) & =  \left(E_M^{-1} - E_A^{-1} \right)\sigma_{MW}(T)\sigma_T + \varepsilon_T\sigma_T, \\
    \label{eq:L2c}
    \Gamma_{TT}(T) & = \left(E_M^{-1} - E_A^{-1} \right)\sigma_T^2.
\end{align}
As recently proved in \cite{Rizzello:2019}, choice (\ref{eq:L3})-(\ref{eq:L2c}) makes the SMA model satisfy energy consistency properties, i.e., it can be represented in a port-Hamiltonian form. This is highly desirable in case a port-based perspective is adopted.

The complete model of the single-crystal SMA wire can be obtained by collecting equations (\ref{eq:MAS3})-(\ref{eq:MAS4a}), (\ref{eq:MAS9}), and (\ref{eq:L3})
\begin{equation} \label{eq:MAS11}
    \begin{cases}
        \dot{\varepsilon} = \dfrac{\textit{v}}{l_0} \\[1em]
        \dot{x}_M = -p_{MA}x_M + p_{AM}(1-x_M) \\[1em]
        \dot{T} = \dfrac{J - \lambda A_s(T-T_E) + L_{x_M}\dot{x}_M}{\Omega\rho_V c_V - L_T} \\[1em]
        f = \pi r_0^2 \dfrac{\varepsilon - \varepsilon_T x_M}{E_M^{-1}x_M + E_A^{-1}(1 - x_M)}
    \end{cases}.
\end{equation}
SMA strain $\varepsilon$, phase fraction $x_M$, and temperature $T$ represent the states of (\ref{eq:MAS11}).
External inputs are the deformation rate $\textit{v}$, the environmental temperature $T_E$, and the Joule heating $J$, while the force $f$ is the system output, cf. Fig.\ref{fig:SMA1}(c). Coupling between SMA and an external load can be performed via the power-conjugated input-output pair $\textit{v}$-$f$.

\subsection{Preliminaries on Hybrid Systems}
\label{sec:Hybrid}
We consider hybrid systems with state $x\in\mathbb{R}^n$ and input $u\in\mathbb{R}^m$ of the form
\begin{equation} \label{eq:HY1}
    \mathcal{H}\colon\left\lbrace
    \begin{array}{ccll}
        \dot{x} & = &f(x, u) & \quad (x, u)\in C\\
        x^+ & \in & G(x) & \quad (x, u)\in D
    \end{array},
    \right.
\end{equation}
where $f\colon\mathbb{R}^{n+m}
\rightarrow\mathbb{R}^n$ is the flow map, $C\subset\mathbb{R}^n$ is the flow set, $D\subset\mathbb{R}^n$ is the jump set, and the set-valued map is  $G\colon\mathbb{R}^n\rightrightarrows\mathbb{R}^n$ the jump map. 
The symbol $\dot{x}$ denotes the time-derivative of state $x$ during flows, while $x^+$ represents the value of state $x$ after an instantaneous change.
To denote the above hybrid system, we use the following shorthand notation $\mathcal{H}=(C, f, D, G)$. A solution pair to $\mathcal{H}$ is any pair $(\phi, u)$, where $\phi$ is a hybrid arc, $u$ is a hybrid signal,  $\dom\phi=\dom u$ that satisfies the dynamics of  $\mathcal{H}$; see \cite{Cai:2009} for formal definitions of hybrid arc, signal, and solution pairs to hybrid systems.
A solution pair is said to be \emph{complete} if its domain is unbounded and \emph{maximal} if it is not the truncation of another solution pair. 
Following \cite{Chai:2021}, we say that $\mathcal{H}$ satisfies the \emph{hybrid basic conditions} if: $C$ and $D$ are closed in $\mathbb{R}^{n+m}$; $f$ is continuous on $C$, $G$ is nonempty, outer semicontinuous\footnote{A set valued map $G\colon\mathbb{R}^{n}\rightrightarrows\mathbb{R}^n$ is outer semicontinuous if its graph is closed; see \cite[Chapter 5]{Goebel:2012}.}, and locally bounded on $D$.
For more details on hybrid systems, the reader may refer to \cite{Goebel:2012}.

\subsection{Hybrid Dynamical Model}
The main result of this paper is presented in this section. The SMA wire model defined in (\ref{eq:MAS11}) provides a good description of the material physical behavior. However, it turns out to be highly stiff from the numerical standpoint. This issue is mainly due to the terms $p_{MA}$ and $p_{AM}$ appearing in state equation \eqref{eq:MAS5}. A potential way to improve the numerical robustness consists of eliminating the stiff dynamics by means of a hybrid reformulation of (\ref{eq:MAS11}).

The key idea is based on a specific structural property of the model. In particular, it can be shown that during phase transformation (i.e. $\dot{x}_M \neq 0$) the following equation provides a tight representation of the dynamics of (\ref{eq:MAS11}) (see \cite{Rizzello:2019})
\begin{equation} \label{eq:L5}
    \begin{cases}
        \sigma(\varepsilon,x_M) = \sigma_A(T) & \text{if } \dot{x}_M > 0 \\
        \sigma(\varepsilon,x_M) = \sigma_M(T) & \text{if } \dot{x}_M < 0 \\
    \end{cases},
\end{equation}
where $\sigma(\varepsilon,x_M)$, $\sigma_A(T)$, and $\sigma_M(T)$ are given by (\ref{eq:MAS3}), (\ref{eq:MAS7}), and (\ref{eq:MAS8}), respectively.
Relationships (\ref{eq:L5}) can be exploited to eliminate the stiff state equation from (\ref{eq:MAS11}), as shown next.

\begin{figure} \vspace{0.1cm}
    \begin{center}
        \includegraphics[width=8.6cm]{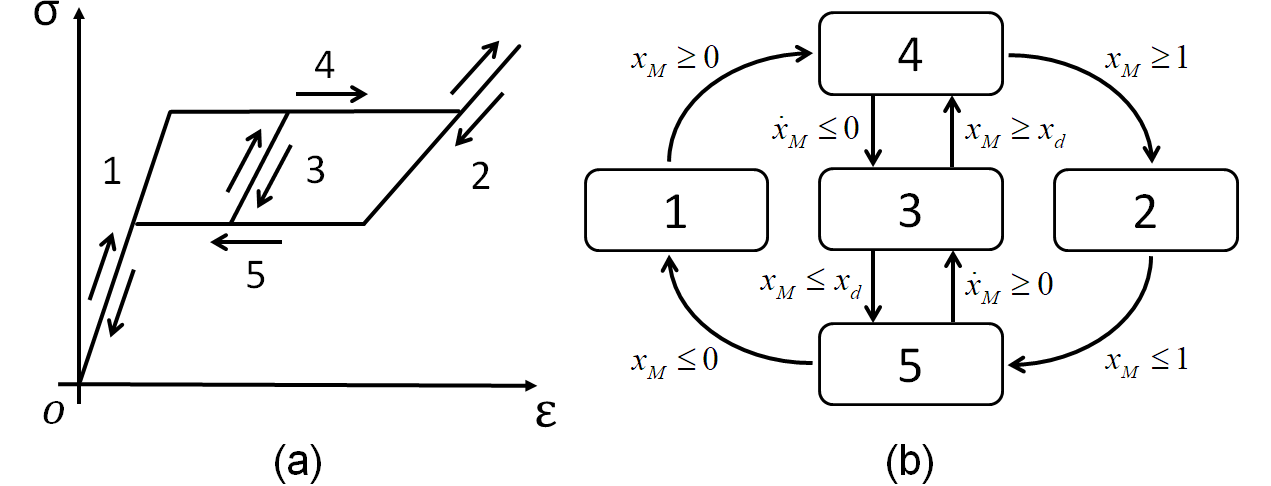}
        \caption{Example of stress-strain hysteresis in single-crystal SMA wire (a) and corresponding hybrid automaton with \emph{modes} and \emph{edges} (b).}
        \label{fig:hyautomaton}
    \end{center}
\end{figure}
By looking at the mechanical hysteretic characteristic of the material reported in Fig.~\ref{fig:hyautomaton}(a), we can identify five different operating modes having the following physical interpretations:
\begin{enumerate}
    \item Full austenitic phase \hfill ($\dot{x}_M = 0$, $x_M = 0$)
    \item Full martensitic phase \hfill ($\dot{x}_M = 0$, $x_M = 1$)
    \item Inner hysteresis loop \hfill ($\dot{x}_M = 0$, $x_M \in (0, 1)$)
    \item Austenite to martensite transformation\hfill ($\dot{x}_M > 0$)
    \item Martensite to austenite transformation\hfill ($\dot{x}_M < 0$)
\end{enumerate}
The transition between those modes can be described via the hybrid automaton in Fig.~\ref{fig:hyautomaton}(b), with the set of \emph{modes} $Q = \{1,2,3,4,5\}$ and the set of \emph{edges} $E = \{(1,4), (2,5), (3,5), (3,4), (4,2), (4,3), (5,1), (5,3)\}$. The mode transition conditions are also reported in Fig.~\ref{fig:hyautomaton}(b).

First, we denote $\dot{\varepsilon}^{(i)}$, $\dot{x}_M^{(i)}$, and $\dot{T}^{(i)}$ the time derivatives of strain, phase fraction, and temperature for the generic mode $i$. For the ease of notation, we also define
\begin{align}
    \label{eq:flux_vareps}
    \dot{\varepsilon}^{(i)} & \coloneqq \phi_{\varepsilon}^{(i)} = 
    \phi_{\varepsilon}^{(i)}(\varepsilon,x_M,T,\textit{v},J,T_E), \\
    \label{eq:flux_xm}
    \dot{x}_M^{(i)} & \coloneqq \phi_{x_M}^{(i)} = 
    \phi_{x_M}^{(i)}(\varepsilon,x_M,T,\textit{v},J,T_E), \\
    \label{eq:flux_T}
    \dot{T}^{(i)} & \coloneqq \phi_{T}^{(i)} =
    \phi_{T}^{(i)}(\varepsilon,x_M,T,\textit{v},J,T_E),
\end{align}
for $i = 1,2,3,4,5$. In order to characterize the hybrid system evolution, auxiliary functions $\phi_{\varepsilon}^{(i)}$, $\phi_{x_M}^{(i)}$, and $\phi_{T}^{(i)}$ need to be properly characterized for each mode. 

It can be readily observed that the state equation for $\varepsilon$ in (\ref{eq:MAS11}) is only affected by the input velocity. Thus, we have
\begin{equation}
    \label{eq:flux_vareps_modes}
    \phi_{\varepsilon}^{(i)} = \dfrac{\textit{v}}{l_0} \qquad \text{with} \quad i = 1,2,3,4,5.
\end{equation}
Based on (\ref{eq:MAS11}), we can also write without loss of generality
\begin{equation}
    \label{eq:flux_T_modes}
    \phi_T^{(i)} = \dfrac{J - \lambda A_s(T-T_E) + L_{x_M}\phi_{x_M}^{(i)}}{\Omega\rho_V c_V - L_T} \;\; \text{with} \;\; i = 1,2,3,4,5.
\end{equation}
Concerning the phase fraction, note that the first three operative modes are characterized by a constant $x_M$, i.e.,
\begin{equation}
    \label{eq:flux_xm_modes123}
    \phi_{x_M}^{(i)} = 0 \qquad \text{with} \quad i = 1,2,3.
\end{equation}
More specifically, we have
\begin{align}
    \label{eq:xm_1_hyb}
    x_M^{(1)} & = 0,\\
    \label{eq:xm_2_hyb}
    x_M^{(2)} & = 1,\\
    \label{eq:xm_3_hyb}
    x_M^{(3)} & \in (0, 1),
\end{align}
where $x_M^{(i)}$ is a shorthand notation to denote the analytical expression of $x_M$ for mode $i$. The actual $x_M^{(3)}$ is defined by the numerical value of $x_M$ prior to the mode change.
By replacing (\ref{eq:flux_xm_modes123}) in (\ref{eq:flux_T_modes}), we obtain the following
\begin{equation}\label{eq:X8}
    \phi_T^{(i)} = \dfrac{J - \lambda A_s(T-T_E)}{\Omega\rho_m c_V - L_T} \qquad \text{with} \quad i = 1,2,3.
\end{equation}
For modes 4 and 5, we can exploit (\ref{eq:L5}) to express $x_M$ as an algebraic function of the other states variables. By replacing (\ref{eq:MAS3}), (\ref{eq:MAS7}), and (\ref{eq:MAS8}) in (\ref{eq:L5}) and solving for $x_M$, we obtain
\begin{align}
    \label{eq:X0}
    x_M^{(4)} &= x_M^{(4)}(\varepsilon,T) = E_M\Sigma_A^{-1}(E_A \varepsilon - \sigma_A(T)), \\
    \label{eq:X1}
    x_M^{(5)} &= x_M^{(5)}(\varepsilon,T) = E_M\Sigma_M^{-1}(E_A \varepsilon - \sigma_M(T)),
\end{align}
where
\begin{align}
    \label{eq:abbr1}
    \Sigma_A &= (E_A - E_M)\sigma_A(T) + E_A E_M \varepsilon_T , \\
    \label{eq:abbr1a}
    \Sigma_M &= (E_A - E_M)\sigma_M(T) + E_A E_M \varepsilon_T \ .
\end{align}
By differentiating (\ref{eq:X0}) and (\ref{eq:X1}) over time and combining it with (\ref{eq:flux_xm})-(\ref{eq:flux_T_modes}), we can solve for $\phi_T^{(i)}$ and thus obtain
\begin{align}
    \label{eq:X6}
    \phi_T^{(4)} &= \dfrac{[J - \lambda A_s(T-T_E)]\Sigma_A^2 + E_A E_M L_{x_M}\Sigma_Al_0^{-1}\textit{v}}{(\Omega\rho_m c_V - L_T)\Sigma_A^2 + E_A E_M L_{x_M}\mathrm{M}}, \\
    \label{eq:X7}
    \phi_T^{(5)} &= \dfrac{[J - \lambda A_s(T-T_E)]\Sigma_M^2 + E_A E_M L_{x_M}\Sigma_Ml_0^{-1}\textit{v}}{(\Omega\rho_m c_V - L_T)\Sigma_M^2 + E_A E_M L_{x_M}\mathrm{M}},
\end{align}
where
\begin{equation}\label{eq:abbr2}
    \mathrm{M} = \left(E_A - E_M\right)\varepsilon\sigma_T + E_M \varepsilon_T\sigma_T .
\end{equation} 
The previous manipulation also leads to the final form of $\phi_{x_M}^{(4)}$ and $\phi_{x_M}^{(5)}$, which is given in compact form as follows
\begin{align} 
    \label{eq:X4}
    \phi_{x_M}^{(4)} &= E_A E_M\Sigma_A^{-2} (\Sigma_A l_0^{-1}\textit{v} - \mathrm{M} \phi_T^{(4)}), \\
    \label{eq:X5}
    \phi_{x_M}^{(5)} &= E_A E_M\Sigma_M^{-2} (\Sigma_M l_0^{-1}\textit{v} - \mathrm{M} \phi_T^{(5)}).
\end{align}
Note that algebraic equations (\ref{eq:xm_1_hyb})-(\ref{eq:xm_3_hyb}), (\ref{eq:X0}), and (\ref{eq:X1}) can be further replaced in (\ref{eq:X8}), (\ref{eq:X6}), and (\ref{eq:X7}), such that $\phi_{\varepsilon}^{(i)}$ and $\phi_{T}^{(i)}$ no longer depend explicitly on $x_M^{(i)}$. This allows us to formally eliminate the phase fraction from the system states. 

Based on the above considerations, we can reformulate model (\ref{eq:MAS11}) as a hybrid system $\mathcal{H}$ in the framework of \cite{Goebel:2012} (cf. Section~\ref{sec:Hybrid}).
In particular, we take as state vector
\begin{equation} \label{eq:X9}
    x\coloneqq \begin{bmatrix} x_1 & x_2 & x_3 & q \end{bmatrix}^\intercal = \begin{bmatrix} \varepsilon & T & x_d & q \end{bmatrix}^\intercal \in \mathbb{X}
\end{equation}
where $\mathbb{X} 
\coloneqq\mathbb{R}_{\geq 0} \times \mathbb{R}_{\geq 0} \times \left[0,1\right] \times \{1,2,3,4,5\}$, and as input vector
\begin{equation}
    u\coloneqq \begin{bmatrix} u_1 & u_2 & u_3 \end{bmatrix}^\intercal = \begin{bmatrix} \textit{v} & J & T_E \end{bmatrix}^\intercal \in \mathbb{U}
\end{equation}
where $\mathbb{U}\coloneqq\mathbb{R} \times \mathbb{R}_{\geq 0} \times \mathbb{R}_{\geq 0}$.
We also define functions $\phi_{\varepsilon}^{(q)}$ and $\phi_T^{(q)}$ according to (\ref{eq:flux_vareps_modes}), (\ref{eq:X8}), (\ref{eq:X6}), and (\ref{eq:X7}) for all $q\in\{1,2,3,4,5\}$.
The main rational behind the hybrid model construction is as follows. The strain $x_1$ and the temperature $x_2$ represent the \emph{continuous states}, which only change during flows. The \emph{frozen phase fraction} $x_3$ computed at the time of mode change, as well as the current mode identifier $q$, define \emph{discrete states} that can only change via jumps. The jump logic, enabling the transitions between two operative modes of the model, is related to the physical interpretation of the material phase transformation, summarized in Fig.~\ref{fig:hyautomaton} (b) (details are omitted for conciseness). 
Hence, the flow set and the flow map can be defined as:
\label{eq:HybridData}
\begin{equation}
\label{eq:hyb_start}
    C \coloneqq \bigcup_{i=1}^{5} C_i \, ,
\end{equation}
where
\begin{footnotesize}
\begin{align} 
    C_1 &\coloneqq 
    \mathbb{R}_{\geq 0} \times \{x\in\mathbb{X}\colon x_M^{(4)} \leq 0\} \times \{0\} \times \{1\}
    ,\\
    C_2 &\coloneqq 
    \mathbb{R}_{\geq 0} \times \{x\in\mathbb{X}\colon x_M^{(5)} \geq 1\} \times \{1\} \times \{2\}
    ,
    \\
    C_3 &\coloneqq
    \mathbb{R}_{\geq 0} \times \{x\in\mathbb{X}\colon x_M^{(5)} \leq x_3 \leq x_M^{(4)}\} \times [0,1] \times \{3\}
    ,
    \\
    C_4 &\coloneqq
       \mathbb{R}_{\geq 0} \times  \{x\in\mathbb{X}\colon x_M^{(4)} \leq 1,\, \phi_{x_M}^{(4)} \geq 0\} \times [0,1] \times \{4\}
    ,
    \\
    C_5 &\coloneqq
        \mathbb{R}_{\geq 0} \times \{x\in\mathbb{X}\colon x_M^{(5)} \geq 0,\, \phi_{x_M}^{(5)} \leq 0\} \times [0,1] \times \{5\}
    ,
\end{align}
\end{footnotesize}
\begin{equation}\label{eq:HY7}
    f(x, u)
    \coloneqq
    \begin{pmatrix}
        \phi_{\varepsilon}^{(q)}, & \phi_T^{(q)}, & 0, & 0 
    \end{pmatrix},\,\,\forall (x, u)\in C
\end{equation}
To enable a jump jump set is defined as:
\begin{equation} \label{eq:HY10}
    D \coloneqq \bigcup_{i=1}^{8} D_i \, ,
\end{equation}
where
\begin{footnotesize}
\begin{align}
    D_1 &\coloneqq 
    \mathbb{R}_{\geq 0} \times \{x\in\mathbb{X}\colon x_M^{(4)} \geq 0,\, \phi_{x_M}^{(4)} \geq 0\} \times \{0\} \times \{1\}
    ,
    \\
    D_2 &\coloneqq
    \mathbb{R}_{\geq 0} \times \{x\in\mathbb{X}\colon x_M^{(5)} \leq 1,\, \phi_{x_M}^{(5)} \leq 0\} \times \{1\} \times \{2\}
    ,
    \\
    D_3 &\coloneqq
    \mathbb{R}_{\geq 0} \times \{x\in\mathbb{X}\colon x_M^{(5)} \leq x_3,\, \phi_{x_M}^{(5)} \leq 0\} \times [0,1] \times \{3\}
    ,
    \\
    D_4 &\coloneqq
    \mathbb{R}_{\geq 0} \times \{x\in\mathbb{X}\colon x_3 \leq x_M^{(4)},\, \phi_{x_M}^{(4)} \geq 0\} \times [0,1] \times \{3\}
    ,
    \\
    D_5 &\coloneqq
    \mathbb{R}_{\geq 0} \times \{x\in\mathbb{X}\colon x_M^{(4)} \geq 1,\, \phi_{x_M}^{(4)} \geq 0\} \times [0,1] \times \{4\}
    ,
    \\
    D_6 &\coloneqq
    \mathbb{R}_{\geq 0} \times \{x\in\mathbb{X}\colon x_M^{(4)} \leq 1,\, \phi_{x_M}^{(4)} \leq 0\} \times [0,1] \times \{4\}
    ,
    \\
    D_7 &\coloneqq 
    \mathbb{R}_{\geq 0} \times \{x\in\mathbb{X}\colon x_M^{(5)} \leq 0,\, \phi_{x_M}^{(5)} \leq 0\} \times [0,1] \times \{5\}
    ,
    \\
    D_8 &\coloneqq
    \mathbb{R}_{\geq 0} \times \{x\in\mathbb{X}\colon x_M^{(5)} \geq 0,\, \phi_{x_M}^{(5)} \geq 0\} \times [0,1] \times \{5\}
    ,
\end{align}
\end{footnotesize}
The jump map is defined so to enforce the transitions in \figurename~\ref{fig:hyautomaton}. In particular, we define:
\begin{equation} \label{eq:HY8}
    G(x) \coloneqq \bigcup_{i\in\{k\in\{1, 2\dots,8\}\colon x\in D_k\}} g_i(x), \,\,\,\,\,x\in D,
\end{equation}
where:
\begin{footnotesize}
\begin{align}
    g_1(x) &\coloneqq  
    \big(
    \begin{matrix}
        x_1, & x_2, & 0, & 4
    \end{matrix}
    \big)&\forall x\in D_1\\
    g_2(x) &\coloneqq 
    \big(
    \begin{matrix}
        x_1, & x_2, & 1, & 5
    \end{matrix}
    \big)&\forall x\in D_2\\
    g_3(x) &\coloneqq 
    \big(
    \begin{matrix}
        x_1, & x_2, & x_3, & 5
    \end{matrix}
    \big)&\forall x\in D_3\\
    g_4(x) &\coloneqq 
    \big(
    \begin{matrix}
        x_1, & x_2, & x_3, & 4
    \end{matrix}
    \big)&\forall x\in D_4\\
    g_5(x) &\coloneqq
    \big(
    \begin{matrix}
        x_1, & x_2, & 1, & 2
    \end{matrix}
    \big)&
    \forall x\in D_5\\
    g_6(x) &\coloneqq  
    \big(
    \begin{matrix}
        x_1, & x_2, & x_M^{(4)}, & 3
    \end{matrix}
    \big)
    &\forall x\in D_6\\
    g_7(x) &\coloneqq 
    \big(
    \begin{matrix}
        x_1, & x_2, & 0, & 1
    \end{matrix}
    \big)&\forall x\in D_7\\
    g_8(x) &\coloneqq\big(
    \begin{matrix}
     x_1, & x_2, & x_M^{(5)}, & 3
    \end{matrix}
    \big)&\forall x\in D_8
    \label{eq:hyb_end}
\end{align}
\end{footnotesize}
Note how explicit dependence of functions $x_M^{(4)}$, $x_M^{(5)}$, $\phi_{x_M}^{(4)}$, $\phi_{x_M}^{(5)}$ on system states and inputs has been omitted from (\ref{eq:hyb_start})-(\ref{eq:hyb_end}) for compactness of notation.

Based on the above defined sets and maps, the single-crystal SMA wire can be modeled via the hybrid system $\mathcal{H}=(C, f, D, G)$.
The result given next establishes a few interesting properties for $\mathcal{H}$.
\begin{proposition}
Let $C$, $f$, $D$, and $G$ be defined as in Section~\ref{sec:3}.B. Then, the following properties hold true for $\mathcal{H}$:
\begin{itemize}
    \item[(a)] $\mathcal{H}$ satisfies the \emph{hybrid basic conditions}; 
     \item[(b)] Let $u$ be a hybrid signal and $\xi\in C\cup D$.  There exists a nontrivial solution pair $(\phi, u)$ to $\mathcal{H}$ such that $\phi(0, 0)=\xi$;
\item[(c)] Let $(\phi, u)$ be any maximal solution pair to $\mathcal{H}$. Then, either $(\phi, u)$ is complete or it has a finite-escape time.
\end{itemize}
\end{proposition}
\begin{remark}
The fact that  $\mathcal{H}$ satisfies the hybrid basic conditions ensures that the proposed model enjoys some robustness properties with respect to small perturbations. 
\end{remark}
\begin{remark}
Having defined $G$ to be set valued 
enables to ensure the satisfaction of the hybrid basic conditions for $\mathcal{H}$; see Section~\ref{sec:Hybrid}. On the other hand, set-valuedness of $G$, along with the closedness of the flow and jump sets,  leads to nonunique solutions. This is a typical consequence when insisting on the satisfaction of the hybrid basic conditions; see, e.g., \cite{Ferrante:2016,Nunez:2016,Ferrante:2017,Ferrante:2018}.    
\end{remark}
\section{Results}
\label{sec:4}

In this section, the new SMA hybrid model is first identified based on available experimental data. Subsequently, comparative simulations between MAS and hybrid models are conducted in the context of a flexible robotic application.

\begin{figure*}[ht!] \vspace{0.1cm}
    \begin{center}
        \includegraphics[width=17cm]{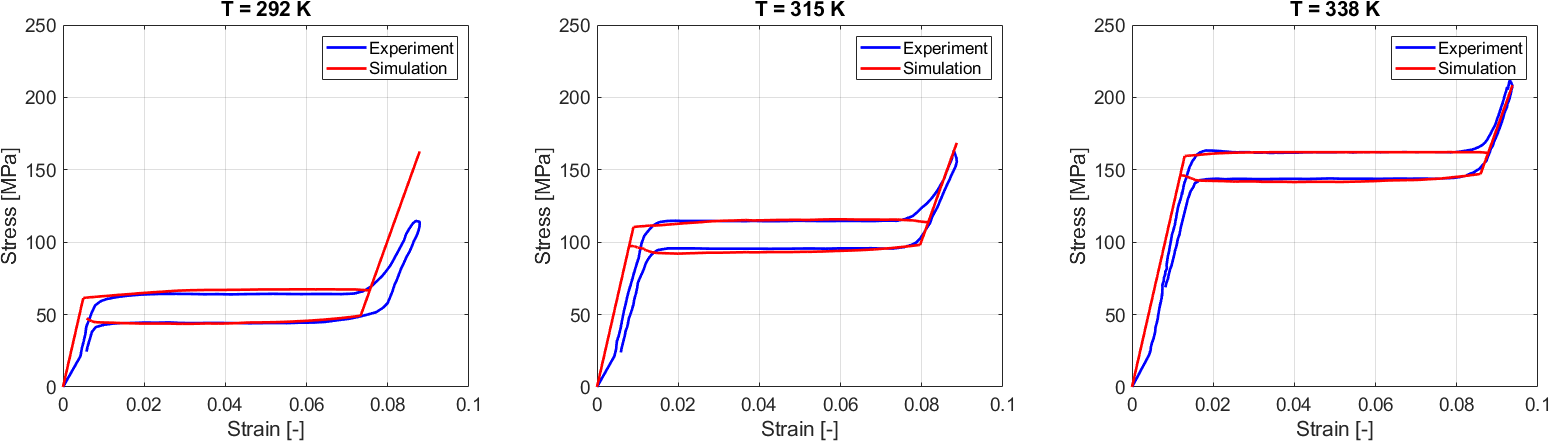}
        \caption{Result of the parameter identification (center) and validation (left- and right-hand sides) based on three different isothermal experiments.}
        \label{fig:ident}
    \end{center}
\end{figure*}

\subsection{Hybrid SMA Model Calibration}
An experimental identification of the SMA hybrid model is first performed, by using the single-crystal material data available in \cite{Fu:1993}. The calibration process is based on an experimental stress-strain measurement, conducted on a CuZnAl single-crystal SMA material working at a constant ambient temperature $T_E = 315$ K. Some of the model parameters can be set \emph{a priori}, i.e., the reference temperature is arbitrarily chosen as $T_0 = 323$ K, while coefficients $r_0 = 75$ $\mu$m, $l_0 = 95.7$ mm, and $\rho_V = 7745$ kg/m$^3$ are derived based on \cite{Fu:1993}. Therefore, the parameters to characterize are: 
\begin{itemize}
    \item Mechanical parameters $E_A$, $E_M$, and $\varepsilon_T$ in (\ref{eq:MAS3});
    \item Thermal parameters $\sigma_T$, $c_V$, and $\lambda$ in (\ref{eq:MAS8a}) and (\ref{eq:MAS9});
    \item Hysteresis parameters $\sigma_{MW}(T_0)$ and $\Delta \sigma$ in (\ref{eq:MAS8a}).
\end{itemize}
The identification, performed via a combination of hand-tuning and nonlinear optimization tools available in \emph{MATLAB}, provides the following optimal values: $E_A = 12.3$ GPa, $E_M = 7.8$ GPa, $\varepsilon_T = 0.067$, $\sigma_T = 2.13$ MPa/K, $c_V = 380$ J/(Kg K), $\lambda = 450$ W/(m$^2$K), $\sigma_{MW}(T_0) = 121.35$ MPa, and $\Delta \sigma = 12.3$ MPa. The resulting curve is shown in Fig.~\ref{fig:ident} (center). In addition, a validation is performed with further stress-strain experiments conducted at $T_E = 292$ K and $T_E = 338$ K, shown in Fig.~\ref{fig:ident} (left-hand side) and Fig.~\ref{fig:ident} (right-hand side), respectively. As it can be seen, the model well predicts the hysteretic curves in each test.

\subsection{Simulation Study: SMA-Driven Flexible Robotic Structure}

\begin{figure}[b!]
    \begin{center}
        \includegraphics[width=8.3cm]{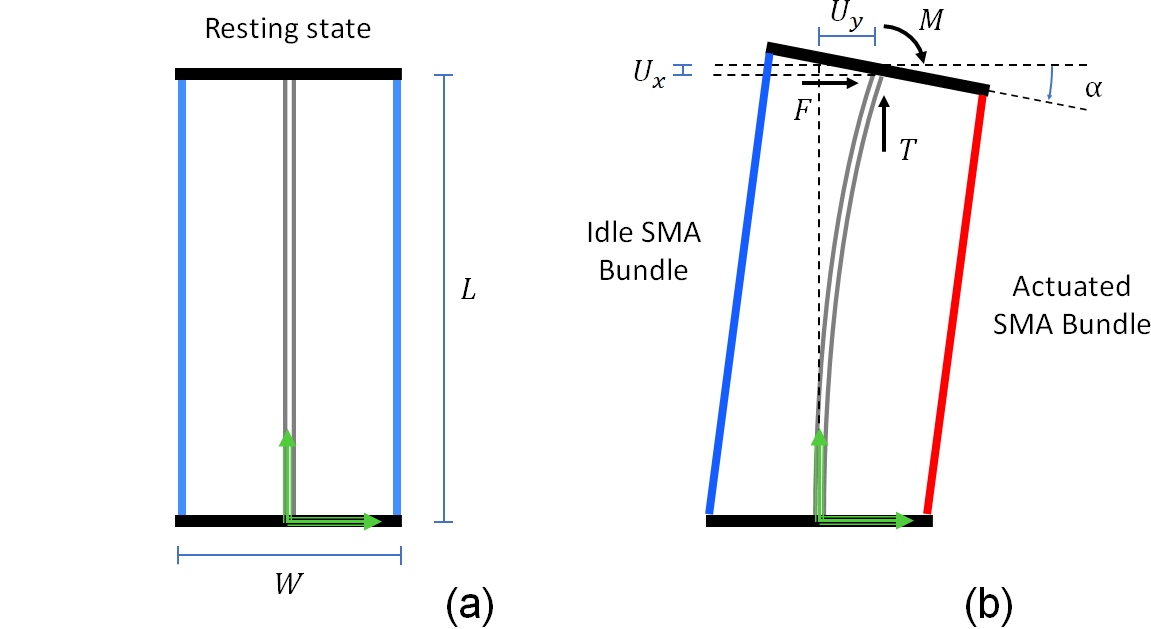}
        \caption{Graphic representation of the flexible robot module actuated by two SMA bundles, idle state (a) and actuated state (b).}
        \label{fig:softrobot1}
    \end{center}
\end{figure}

The aim of this section is to compare the performance of MAS and hybrid models in terms of both accuracy and simulation time, based on numerical studies conducted on a SMA-actuated device. In order to test the SMA hybrid model within a meaningful setting for continuum robotic applications, the multi-actuated structure shown in Fig.~\ref{fig:softrobot1} is considered as numerical case study. It consists of a T-shaped backbone made of a flexible cylindrical beam and a rigid top plate. Two tendon-like actuators made of bundles of pre-tensioned SMA wires are mounted in parallel to the beam. 
The considered system operates similarly to the flexible module described in Section~\ref{sec:2}, with the only difference that in here the motion is constrained on a plane, for simplicity.
With respect to Fig.~\ref{fig:softrobot1}, the size of both rigid plate and base is denoted as $W$. The flexible backbone is modeled as a planar Euler-Bernoulli beam \cite{Gere:1997} (small deformation assumption). In this way, the beam can be uniquely described by three spatial degrees of freedom, denoted as $q = [U_x\;U_y\;\alpha]^\intercal$, which represent the displacement and inclination of the tip, see Fig.~\ref{fig:softrobot1}(b). The corresponding energy-conjugated generalized forces applied to the beam by the SMAs are denoted as $\tau = [T\;F\;M]^\intercal$, and are reported in Fig.~\ref{fig:softrobot1}(b) as well. We also assume that the mass of the rigid plate dominates the one of the beam. Based on the above assumptions, the following model is obtained
\begin{align}
    \label{eq:EB-Ba}
    m_H\ddot{U}_x &= - EAL^{-1}U_x - b_x\dot{U}_x +T, \\
    \label{eq:EB-Bb}
    m_H\ddot{U}_y &= -12EIL^{-3}U_y + 6EIL^{-2}\alpha - b_y\dot{U}_y + F, \\
    \label{eq:EB-Bc}
    J_H\ddot{\alpha} &= 6EIL^{-2}U_y - 4EIL^{-1}\alpha - b_{\alpha}\dot{U}_{\alpha} + M.
\end{align}
where $E$ is the Young's modulus of the beam, $A$ is the cross-sectional area of the beam, $L$ is the beam length, $I$ is the second moment of area of the beam, while $m_H$ and $J_H$ are the mass and the moment of inertia of the top platform, respectively. Dissipative phenomena are rendered through viscous friction coefficients $b_x$, $b_y$, and $b_{\alpha}$, while gravitational effects are neglected for simplicity.
To establish a physically consistent coupling between the flexible structure and the SMA actuators, we define the following power-preserving interconnection
\begin{equation}
    \label{eq:EB-B1}
    \begin{bmatrix}
        \textit{v}\\
        \tau
    \end{bmatrix}
    = 
    \begin{bmatrix}
        0 & J(q) \\
        -J^\intercal(q) & 0
    \end{bmatrix}
    \begin{bmatrix}
        f\\
        \dot{q}
    \end{bmatrix},
\end{equation}
where $f$ and $\textit{v}$ denote (with an abuse of notation) the vectors of SMA forces and velocities, linked together via the model developed in Section~\ref{sec:3}, while $J(q)$ is the Jacobian matrix which relates the generalized velocities $\dot{q}$ to the SMA velocities $\textit{v}$. Such a matrix is derived via the kinematic model
\begin{equation}
    \label{eq:EB-B3}
    J(q) := \frac{\partial l(q)}{\partial q},
\end{equation}
where the vector of SMA lengths $l$ is simply given by
\begin{small}
\begin{equation}
    \label{eq:EB-B2a}
    l(q) = \left[ \begin{array}{c}
        \sqrt{[U_y-\frac{W}{2}(1-\cos\alpha)]^2+(L+U_x-\frac{W}{2}\sin\alpha)^2}  \\
         \sqrt{[U_y+\frac{W}{2}(1-\cos\alpha)]^2+(L+U_x+\frac{W}{2}\sin\alpha)^2}
    \end{array} \right].
\end{equation}
\end{small}

The overall system model, provided with the hybrid system $\mathcal{H}$, is implemented in \emph{MATLAB}/\emph{Simulink} environment via the \emph{Hybrid Equation (HyEQ) Toolbox} \cite{Sanfelice:2013}. 
\begin{remark}
As mentioned in \textit{Remark 2}, solutions of system $\mathcal{H}$ are nonunique. Nonuniqueness can be handled in simulations by setting specific rules, thereby performing a selection of the most suitable solution.
\end{remark}

The control input of each SMA actuator is an electrical power, thus it must always be greater or equal than zero.
We denote as $J_1$ and $J_2$ the Joule heating signals used to control SMA bundle 1 and 2, respectively.
For simplicity, we assume that the two bundles are never activated simultaneously.
To implement this driving strategy in a compact way, we define a new virtual command $J_{eq}$ which can be either positive or negative. 
Signal $J_{eq}$ uniquely determines both $J_1$ and $J_2$ according to the following rules: 
\begin{itemize}
\item If $J_{eq} > 0$ then set $J_1 = |J_{eq}|$ and $J_2 = 0$, i.e., only the first wire bundle is activated; 
\item If $J_{eq} < 0$ then set $J_1 = 0$ and $J_2 = |J_{eq}|$, i.e., only the second wire bundle is activated;
\item If $J_{eq} = 0$ then set $J_1 = 0$ and $J_2 = 0$, i.e., both wire bundles are not activated;
\end{itemize}
Therefore, the sign of the new input signal $J_{eq}$ determines which bundle is activated, while the magnitude of $J_{eq}$ determines the amount of corresponding Joule heating.

\begin{figure}
    \begin{center}
        \includegraphics[width=8.5cm]{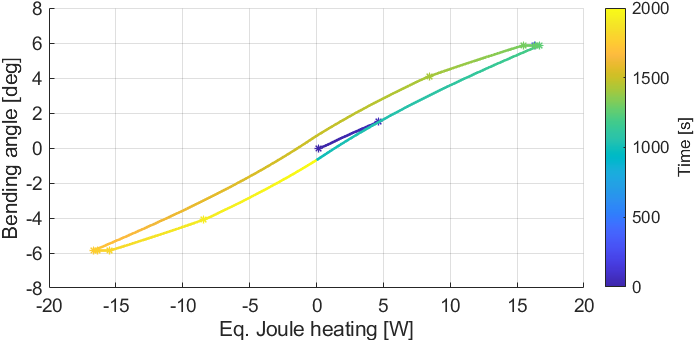}
        \caption{Input-output hysteresis of the system, 1 mHz sinusoidal input.}
        \label{fig:plotsin}
    \end{center}
\end{figure}

All simulations are conducted by considering the following parameters, chosen in an arbitrary yet realistic way: $W = 10$ mm, $L = 100$ mm, $E = 2$ GPa, $m_H = 10$ g, $J_H = \sfrac{W^2m_H}{12}$, $h = 2.5$ mm, $I = \sfrac{h^4\pi}{4}$, $T_E = 298$ K, $b_x = b_y = $ 2 N$\cdot$s/m and $b_\alpha = $ 2 N$\cdot$m$\cdot$s. To simulate the MAS model, the additional parameters appearing in (\ref{eq:MAS6}) are given as $\omega_x = 100$ Hz and $V_L = 5 \cdot 10^{-23}$ m$^{3}$, as in \cite{Furst:2012}. Finally, a number of $n = 10$ SMA wires is considered in each bundle, implying that the SMA force and power of the single-wire model must be scaled by a factor $n$ and $1/n$, respectively. For each simulation, the solver \emph{ode15s} is chosen to deal with the high stiffness of the beam model.
\begin{figure}  \vspace{0.1cm}
    \begin{center}
        \includegraphics[width=8cm]{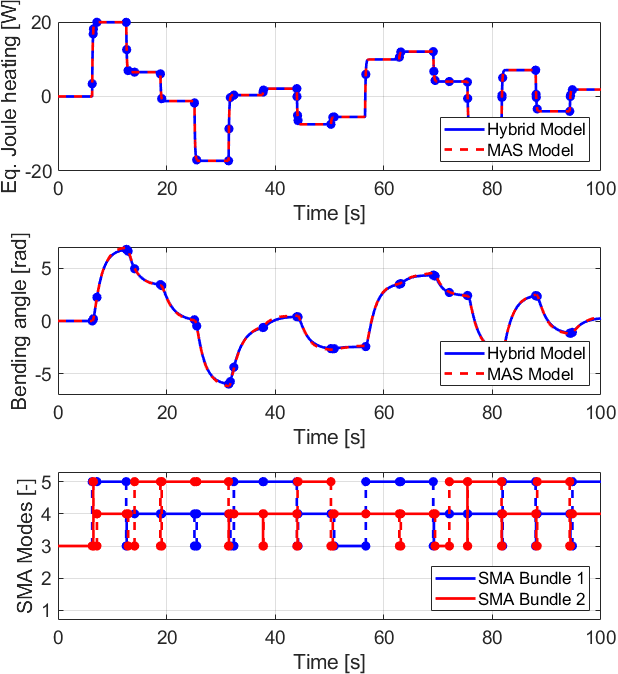}
        \caption{Comparison between the hybrid dynamical model (solid blue line) and  MAS model (dashed red line) with a random input.}
        \label{fig:plotrand}
    \end{center}
\end{figure}
A comparative simulation campaign is conducted, in which both hybrid and MAS models are compared when activated with the same control input $J_{eq}$. A number of 30 different inputs are randomly generated, chosen as sequences of steps with random amplitudes and duration. Each input signal has a total duration of 100 s. One example of such a signal is shown Fig.~\ref{fig:plotrand}. The same figure also reports the output of both models in terms of $\alpha$, as well as the jumps that occur for both SMA wires in case of the hybrid implementation.
As it can be observed, the response curves of both models are practically indistinguishable. This fact holds true for all the conducted simulations, thus confirming the validity of the hybrid implementation.
The simulation time of the hybrid model, however, is remarkably smaller than the MAS one (1.75 s vs. 8.66 s, averaged over the 30 simulations). This result allows us to asses the improved numerical properties of the new model, at least for the given class of robotic systems.
Finally, Fig.~\ref{fig:plotsin} shows the angle $\alpha$ as a function of $J_{eq}$, for the hybrid model only, in case of a 1 mHz sinusoidal input. The plot clearly shows the input-output hysteretic behavior of the system.

\section{Conclusion}
\label{sec:5}

In this paper, a hybrid description of the hysteresis occurring in single-crystal SMA wires is presented, based on a reformulation of the MAS model. The adopted port-oriented representation provides a ready-to-use modeling framework that supports simulation, optimization, and control of SMA-driven robotic structures. The obtained results show how the hybrid model presents a remarkable accuracy in describing the system dynamics, while requiring only 20\% of the time needed to simulate the MAS model. Such a reduction in computation time will play a key role when simulating complex continuum robots, actuated by a significantly larger number of SMAs. 

Future developments will concern the extension of the hybrid reformulation to commercially available SMA materials (i.e., NiTi) which exhibit a so-called polycrystalline (rather than single-crystal) behavior. A polycrystalline material is characterized by several inhomogeneities and impurities at mesoscopic level. This results into a more complex hysteresis curve, characterized by a smooth shape and multiple inner loops, whose accurate modeling turns out to be highly challenging. An experimental test bench for characterizing continuum robotic structures will also be assembled, and used to evaluate the performance of the hybrid model in fully describing real-life SMA actuated systems. Control and self-sensing algorithms based on the hybrid framework will also be developed.
\balance
\bibliographystyle{IEEEtran}
\bibliography{IEEEref}

\end{document}